\documentclass[letterpaper, 10 pt, conference]{ieeeconf}  

\usepackage[final]{graphicx}
\usepackage{amsmath}
\usepackage{float}
\usepackage{algorithm}
\usepackage[noend]{algpseudocode}
\usepackage{fontawesome5}
\usepackage[symbol]{footmisc}
\usepackage{subfigure}
\usepackage{xcolor}
\usepackage{titlesec}
\usepackage{hyperref}

\renewcommand{\thesubsubsection}{\arabic{subsubsection}) }
\titleformat{\subsubsection}[runin]{\bfseries}{\thesubsubsection}{0em}{}
\titlespacing{\subsubsection}{0pt}{0.5ex}{1ex}

\newcommand{\subsubsubsection}[1]{\textit{#1}}
\newcommand{\subsubsectioncol}[1]{\subsubsection{#1:}}

\newcommand{\policysymbol}{\triangle}
\newcommand{\policy}{$\pi^{\policysymbol}$}
\newcommand{\policyR}{$\pi^{\policysymbol}_{R}$}
\newcommand{\policyP}{$\pi^{\policysymbol}_{P}$}
\newcommand{\targetedpolicy}{$\pi^{T}_{R}$}
\newcommand{\shortname}{MAGIC}
\newcommand{\cleanfullpolicyname}{Morphology-Agnostic Controller }
\newcommand{\fullpolicyname}{\underline{M}orphology-\underline{AG}nost\underline{I}c \underline{C}ontroller }
\newcommand{\fullpolicynamenobold}{Morphology-AGnostIc Controller }

\newcommand{\ourmorph}{\textit{MAGIC-optimized}}

\newcommand{\morphsymbol}{\star}
\newcommand{\mours}{$m^{\morphsymbol}$}

\newcommand{\humansymbol}{\mathcal{H}}
\newcommand{\mhuman}{$m^{\humansymbol}$}

\newcommand{\onestagehalfname}{privileged single-stage learning }
\newcommand{\onestagefullnamecased}{\underline{PRI}vileged \underline{S}ingle-stage learning via latent align\underline{M}ent }
\newcommand{\onestagefullname}{PRIvileged Single-stage learning via latent alignMent }

\newcommand{\onestageshortname}{PRISM}

\definecolor{stateblue}{HTML}{000000}
\newcommand{\stateblue}[1]{\textcolor{stateblue}{#1}}
\definecolor{statered}{HTML}{000000}
\newcommand{\statered}[1]{\textcolor{statered}{#1}}
\definecolor{stateyellow}{HTML}{000000}
\newcommand{\stateyellow}[1]{\textcolor{stateyellow}{#1}}
\definecolor{stategreen}{HTML}{000000}
\newcommand{\stategreen}[1]{\textcolor{stategreen}{#1}}
\definecolor{stateorange}{HTML}{000000}
\newcommand{\stateorange}[1]{\textcolor{stateorange}{#1}}

\newcommand{\colorobs}{\stateblue{\mathbf{o_t}}}
\newcommand{\colorstate}{\statered{\mathbf{s_t}}}
\newcommand{\colormorph}{\stateyellow{\mathbf{m_i}}}

\newcommand{\colorzR}{\stateblue{\mathbf{z^R}}}
\newcommand{\colorzP}{\statered{\mathbf{z^P}}}
\newcommand{\colorzM}{\stateyellow{\mathbf{z^M}}}

\newcommand{\coloractP}{\stateblue{\mathbf{a^P}}}
\newcommand{\coloractR}{\statered{\mathbf{a^R}}}

\IEEEoverridecommandlockouts                              

\title{\LARGE \bf
On Designing a Learning Robot: Improving Morphology\\ for Enhanced Task Performance and Learning
\thanks{\hspace{-3mm}Correspondence to: {\tt\small maks@gatech.edu}}
\thanks{\hspace{-3mm}* The research was conducted during Residency at Everyday Robots.}
\thanks{\hspace{-3mm}$^{1}$Everyday Robots,$^{2}$ Georgia Institute of Technology,$^{3}$Robotics at Google}
\thanks{\hspace{-3mm}$^{4}$Stanford University, $^{5}$Work done while at Everyday Robots}
}

\author{
    Maks Sorokin$^{1*,2}$,
    Chuyuan Fu$^{1}$,
    Jie Tan$^{3}$,
    C. Karen Liu$^{4}$,
    \\
    Yunfei Bai$^{5}$,
    Wenlong Lu$^{1}$,
    Sehoon Ha$^{2}$, 
    Mohi Khansari$^{1}$
}
\begin{document}

\maketitle
\thispagestyle{empty}
\pagestyle{empty}

\begin{abstract}

As robots become more prevalent, optimizing their design for better performance and efficiency is becoming increasingly important.
However, current robot design practices overlook the impact of perception and design choices on a robot's learning capabilities.
To address this gap, we propose a comprehensive methodology that accounts for the interplay between the robot's perception, hardware characteristics, and task requirements.
Our approach optimizes the robot's morphology holistically, leading to improved learning and task execution proficiency. 
To achieve this, we introduce a Morphology-AGnostIc Controller (MAGIC), which helps with the rapid assessment of different robot designs.
The MAGIC policy is efficiently trained through a novel PRIvileged Single-stage learning via latent alignMent (PRISM) framework, which also encourages behaviors that are typical of robot onboard observation.
Our simulation-based results demonstrate that morphologies optimized holistically improve the robot performance by \mbox{15-20\% on various} manipulation tasks, and require 25x less data to match human-expert made morphology performance.
In summary, our work contributes to the growing trend of learning-based approaches in robotics and emphasizes the potential in designing robots that facilitate better learning. The project's website can be found at \href{https://learning-robot.github.io}{learning-robot.github.io}

\end{abstract}



\section{INTRODUCTION}

\looseness=-1
Recent advances in hardware and software make autonomous robots more and more important in various environments, from manufacturing and warehouses to healthcare and living spaces.
Learning has been one of the most promising tools for operating robots in such unstructured environments, enabling them to acquire complex perception and reasoning capabilities.
However, the current status quo of designing robots does not account for the impact of learning: rather, many robots are still designed based on human experts' intuition or hand-crafted heuristics.
Therefore, such designs can lead to a sub-optimal performance by causing unexpected visual occlusions.
This is where the idea of guiding robot design to improve the robot learning capability comes in, inspired by the evolutionary process.

\looseness=-1
The evolution of physical attributes through natural selection has played a significant role in the emergence of advanced cognitive abilities among living beings~\cite{manning1998introduction}.
By embracing the idea of evolution, robots also have the potential to evolve their designs for better real-world performance. 
However, it is extremely challenging to encapsulate all the perception, control, and hardware design into a single holistic evaluation pipeline due to the complexity of the components. 
For instance, it will be extremely expensive to formulate a typical two-loop design optimization process, which searches robot morphologies in the outer loop and trains a policy for each given design in the inner loop.
As such, traditional design optimization techniques focus on enhancing certain attributes in isolation or exploring based on hand-designed heuristics.

\begin{figure}
  \centering
  \vspace{3mm}
  \includegraphics[width=0.5\textwidth]{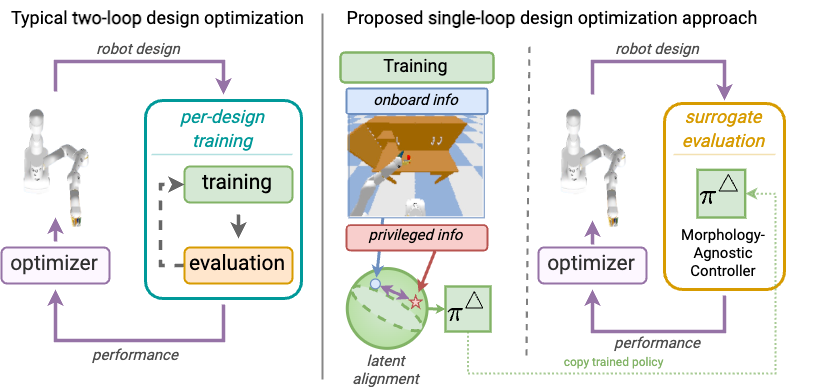}
  \vspace{-6mm}
  \caption{
  A side-by-side comparison of existing vs proposed hardware design optimization approaches.
  }
  \vspace{-5mm}
  \label{fig:teaser}
\end{figure}

\looseness=-1
This work particularly aims to discuss the design optimization for vision-based manipulation.
In the general context of manipulation, visual sensors provide a rich stream of information that allow robots to perform tasks such as grasping, object manipulation, and assembly.
The use of visual sensors in manipulation, however, inevitably poses challenges associated with complete or partial field-of-view occlusion, which can degrade performance due to limited perception. 
Whether an occlusion is harmless or fatal often depends on the task at hand and the stage of task execution.
Such scenarios motivate us to explore hardware optimization without neglecting the interplay between the robot's morphology, onboard perception abilities, and their interaction in different tasks.

\looseness=-1
We present a learning-oriented morphology optimization framework to improve the initial robot design crafted by a human expert.
Our key idea is to develop a \fullpolicynamenobold (\shortname) that is capable of controlling a range of morphologies using onboard visual observations, greatly reducing the costs of traditional two-loop design optimization. 
The controller is trained using a novel \onestagefullname (\onestageshortname) formulation, which unifies the typical two-stage approach of teacher and student training \cite{chen2020learning}.
In our experience, this novel formulation is essential for the student policy to learn behaviors characteristic of its own observation capability and not of a privileged agent.
Once the morphology-agnostic controller is learned, we optimize robot design parameters using Vizier~\cite{google_vizier} optimizer using the controller's performance as a surrogate measure (Fig. \ref{fig:teaser}).
We leverage simulation throughout this work to accelerate the process of looking for an optimal configuration without actually building a physical robot, which is both costly and time-consuming.

\looseness=-1
We evaluate the proposed technique for optimizing the morphology of a mobile manipulator.
We find that our framework can find an improved morphology that shows better performance on overall tasks and facilitates a more sample efficient learning.
Specifically, an optimized design leads to a $15$-$20$\% success rate improvement on various manipulation tasks and is $25$x more data efficient when trained from scratch.
With this work, we would like to highlight the untapped potential of learning-based robot design optimization and show how robot designs can be tailored for better performance and learning with onboard sensing.
\section{RELATED WORK}

\subsection{Morphology Optimization}

\looseness=-1
Robot's physical design is an important factor in its performance and ability to complete assigned tasks.
Traditionally, robot designs are made by human experts who rely on heuristics to optimize the structure.
A number of approaches have been developed to algorithmically assist designers in making better physical structures~\cite{coello1998using, kouritem2022multi}, improving reachability~\cite{seraji1995reachability, vahrenkamp2009humanoid, makhal2018reuleaux}, and observability of the workspace~\cite{triggs1995automatic, chen2004automatic, nikolaidis2009optimal}.
However, heuristic approaches don't offer any insight on the design's performance on actual tasks.
This is particularly true for tasks that involve vision and learning, where the robot's physical structure can significantly impact learning efficiency and performance. 

\looseness=-1
Recently, researchers have aimed to address the limitations of robot designs by optimizing their physical structures in an informed and systematic way.
A number of works have proposed computational approaches for co-optimizing the design parameters and motion trajectories of robotic systems~\cite{ha2018computational, ha2018computational2}.
Typically these works model the relationship between form and function as solutions to an optimal control problem, often showing that such frameworks are effective in optimizing robotic design for a variety of tasks.
Recently, a number of methods were proposed for the joint optimization of physical structures using policy optimization methods~\cite{wampler2009optimal, ha2019reinforcement, zhao2020robogrammar}.
While these approaches bring us closer to optimizing the designs for performance on the actual tasks, so far they tackle cases with proprioceptive sensor information and do not take into account vision, which is often critical in deep learning approaches~\cite{saycan2022arxiv, rt12022arxiv}.
In this work, we incorporate exteroceptive information obtained from the onboard camera sensor to account for an interplay between the design and the robot's ability to sense task-relevant information.

\subsection{Vision-based Robot Learning}

\looseness=-1
Visual sensing plays a crucial role in the field of robotics, enabling the robots' perception and understanding of their surroundings. 
The field has come a long way in terms of explicitly understanding the world through pixels ~\cite{he2016deep, He_2017_ICCV, dosovitskiy2020image} and 3D ~\cite{monodepth2, philion2020lift, guizilini20203d}, as well as, implicitly through learned features that enhance downstream task performance~\cite{grill2020bootstrap, shah2021rrl}.

\looseness=-1
Utilizing the advances of learning, the field of robotics has been booming with pipelines that leverage Behavioral Cloning~\cite{pomerleau1988alvinn} and Reinforcement Learning~\cite{sutton2018reinforcement} methods.
Many systems that are designed to be deployed autonomously in the real-world leverage both exteroceptive sensing and learning.
Notably, a number of projects that were deployed in the real-world leveraged learning and vision, some examples of such systems include robot locomotion~\cite{yu2021visual, miki2022learning, agarwallegged}, navigation~\cite{tolani2021visual, hoeller2021learning, sorokin2022learning}, and manipulation~\cite{kalashnikov2018qt, xia2020relmogen, jang2022bc, khansari2022practical}.
These works focus on improvements to the training pipelines and algorithms while keeping the robot characteristics intact.
In this work, we leverage findings from vision-based robotics research and use manipulation problems as a testing ground for design optimization.

\subsection{Surrogate Evaluation}

\looseness=-1
A surrogate function is a model that approximates the behavior of the true function that is costly to produce~\cite{sobester2008engineering}.
In the context of robot design optimization, a true function would measure the quality of the design through a full cycle of data collection, training, and evaluation.
Alas, with a large number of designs to rate, it is infeasible to go through the whole process from collection to evaluation for each design.
Thus, we explore the idea of a morphology-agnostic controller~\cite{Yu-RSS-17,zhao2020robogrammar,gupta2022metamorph} which is capable of controlling a range of morphologies.
Zhao~et~al.~\cite{zhao2020robogrammar} used a model predictive control to evolve terrain traversal creatures. 
Such controllers can be used as a \textit{proxy} to evaluate the performance of actual robot designs.

\subsection{Universal Policy}
\looseness=-1
A universal policy can be deﬁned as a controller which is able to operate in different environments while performing various tasks.
In the context of learning, universal controllers are made possible through the exposure of the controller to a diverse set of environments during training~\cite{tobin2017domain}.
Yu~et~al.~\cite{Yu-RSS-17} showed that it is possible to effectively adapt the controller to unknown task environments.
Gupta~et~al.~\cite{gupta2022metamorph} leveraged learning and demonstrated cross-morphology policy transfer in the context of locomotion.
In~RT-1~\cite{rt12022arxiv}, it was recently shown how a large-scale training could enable effective multi-task learning in different environments and help with cross-robot transfer.
These approaches provide a strong foundation to enable generalization to unseen environments and tasks.
In this work, we propose a recipe to enable the training of a universal controller that works across different morphologies while using the robot's \textit{onboard} sensing.

\begin{figure*}[!ht]
  \centering
  \includegraphics[width=1.0\textwidth]{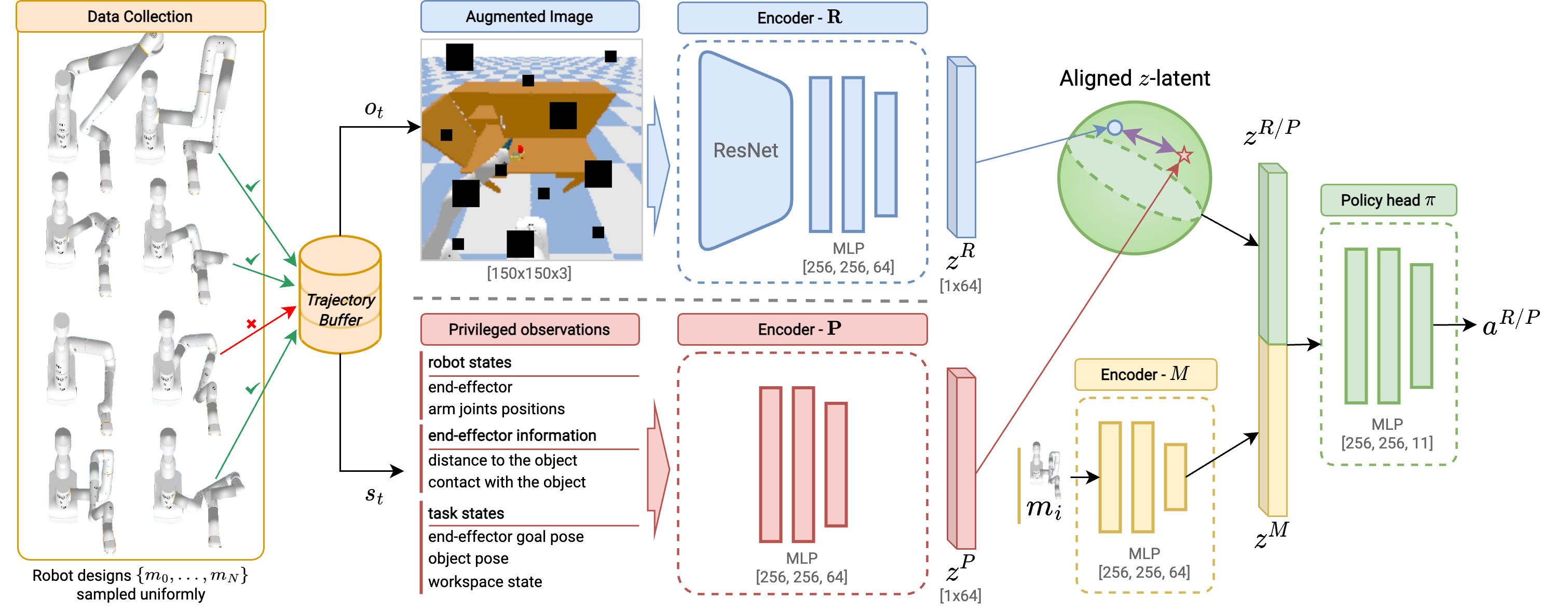}
  \vspace{-7mm}
  \caption{
  \textbf{Overview of policy \policy\ a \cleanfullpolicyname trained via \onestagefullname (\onestageshortname)}
  \protect\\
  \textit{\stateorange{Data Collection}} - demonstrates a few robot designs randomly generated during the collection process. Robot trajectories are collected using motion planning and only successful trajectories are stored in the Trajectory Buffer.
  \textit{\stateblue{Encoder $R$}} processes information obtained from onboard sensors (e.g., cameras) to generate embedding $\colorzR$.
  \textit{\statered{Encoder $P$}} extracts $\colorzP$ from privileged state which contains a comprehensive information about the robot and workspace.
  $\colorzP$ and $\colorzR$ are explicitly aligned through a loss function to produce mutual information.
  Then, \textit{\stategreen{policy head $\pi$}} generates onboard and privileged actions $\coloractP$/ $\coloractR$ using morphology embedding $\colorzM$, and $\colorzR$/$\colorzP$.
  As such onboard (\policyP$\gets\pi(\colorzP,\colorzM$)) and privileged (\policyR$\gets\pi(\colorzR,\colorzM$)) policies are trained jointly via latent alignment and share \textit{\stategreen{policy head $\pi$}}. For details refer to Section \ref{sec:universal_pi}.
  }
  \vspace{-4mm}
  \label{fig:architecture}
\end{figure*}

\looseness=-1
To better guide the controller during training, we leverage ideas of privileged learning~\cite{chen2020learning} to bootstrap the training.
A privileged pipeline enables a more capable and sample efficient policy training that can control different morphologies.
Overall, using the morphology-agnostic controller as a surrogate function allows us to efficiently explore the design candidate space and find an improved morphology design.

\section{\cleanfullpolicyname}\label{sec:universal_pi}

\subsection{Overview}
We present a \fullpolicyname (\shortname) capable of operating across a variety of environments, tasks, and robot morphologies.
The controller is a neural network policy \policy\ trained to control a wide range of robot morphologies using the robot's onboard camera sensing.
To achieve this challenging goal, we introduce a \onestagehalfname framework (detailed in Section \ref{subsec:training}) to bootstrap the behavioral learning from demonstrations.
The data is collected across a spectrum of robot designs (detailed in Section \ref{sec:exp_setup}) to make the policy compatible with morphological variations.
The \policy\ training pipeline and policy architecture are summarized in Fig. \ref{fig:architecture}.
The \policy\ will be later used as a surrogate function to rapidly assess the quality of different morphologies in Section \ref{sec:hw_optim}.
Below, we introduce the policy structure and training protocols used for obtaining such a policy.

\subsection{States and Observation}
We train our policy using Behavior Cloning, which has been proven effective for a variety of manipulation tasks~\cite{jang2022bc, khansari2022practical, rt12022arxiv}. In this paper, we use the following notation to describe the states and observations: \stateblue{\textit{onboard}} observations~${o_t}$, \statered{\textit{privileged}} states ${s_t}$, \stateyellow{\textit{morphology}} configuration ${m_i}$.
\stateblue{\textit{Onboard}} observations consist of sensor readings which are available during robot deployment. \statered{\textit{Privileged}} states provide a comprehensive representation that is only used to bootstrap the learning efficiency of the policy. These states do not suffer from sensor limitations or occlusions, notably, they are unobtainable during the policy deployment. \stateyellow{\textit{Morphology}} configuration is used to inform the policy about the robot design parameters.

States and observations used in our tasks of robot manipulation are the following:
\begin{itemize}
    \item \stateblue{$I \in R^{[150\times150\times3]}$} (onboard) - image observations rendered from the robot's onboard camera.
    \item \statered{$J \in R^{8}$} (privileged) - positions of the robot's arm joints.
    \item \statered{$E^{curr/goal} \in R^{11}$} (privileged)  - current and goal states of the robot's end-effector, including Cartesian coordinates, quaternion orientation, and finger positions.
    \item \statered{$B \in R^{7}$} (privileged) - position and orientation of the object being manipulated during the task.
    \item \statered{$F \in R^{2}$}  (privileged) - an indicator of contact between the fingers and the object, and the distance between them.
    \item \statered{$W \in R^{2}$}  (privileged) - openness state of the cabinet, used to track the progress of the closing task.
    \item \stateyellow{$m \in R^{7}$} (morphology)- configuration parameters that define the robot's physical structure and configuration.
\end{itemize}
These observations and states provide a comprehensive understanding of the robot's environment and actions, which are used to train the \policy.

The policy output controls the displacement of robot end-effector position, orientation, and two-fingers.

\subsection{Privileged Single-stage Learning via Latent Alignment}\label{subsec:training}
In this section, we describe the \onestagefullnamecased (\onestageshortname) used for training of~\policy.
Privileged learning has been shown to be effective for a number of applications~\cite{chen2020learning, sorokin2022learning, miki2022learning, agarwallegged}.
Generally in this paradigm, two policies are trained sequentially, a student policy with limited observations, and a privileged policy with full observations.
The privileged policy's sole role is to guide the student during the transfer process, leading to improved learning efficiency and improved behavior due to a stronger learning signal.

We propose a novel training procedure, which unifies the traditional two-stage approach into one stage by using a latent space alignment loss during the optimization process.
This unification allows us to discourage the student policy from learning \textit{supernatural} behaviors which incorrectly exploit the information only present in privileged states, which is important to learn realistic vision-based policies.
To make this approach possible, we use a combination of loss functions to train the policy network.
We use Behavioral Cloning (BC) for action and Latent Alignment loss for the alignment of information in the encoder latent space.
We utilize this framework for training \policy\ that is agnostic of the robot morphology in both observation and control space.

\vspace{1mm}
\subsubsectioncol{Network Architecture} Our network architecture consists of four main components: image encoder \stateblue{$R$}, a privileged encoder \statered{$P$}, morphology encoder \stateyellow{$M$}, and the policy head~\stategreen{$\pi$}.
Encoders produce unit-vector embeddings
$\colorzR$,
$\colorzP$,
$\colorzM$
processing the respective inputs:
$\colorobs$\footnote[1]{For reaching $o_t$ consists on $E^{goal}$ in addition to $I$},
$\colorstate$, and
$\colormorph$.
Onboard policy action $\coloractR$ and privileged policy action $\coloractR$ are produced by querying policy head with corresponding inputs:
$\stategreen{\pi}(\colorzR,\colorzM)$ and
$\stategreen{\pi}(\colorzP,\colorzM)$.
For the rest of the paper, we refer to
onboard policy $\stategreen{\pi}(\colorzR,\colorzM)$ as \policyR, and to
privileged policy $\stategreen{\pi}(\colorzP,\colorzM)$ as \policyP.
For a detailed overview of the architecture refer to Fig. \ref{fig:architecture}.

\vspace{1mm}
\subsubsectioncol{Losses} We use a combination of three loss function terms for training the policy network: two behavioral cloning terms for actions and one latent alignment term for the alignment of information in the encoder latent space.

\vspace{1mm}
\subsubsubsection{Behavioral Cloning loss.} $L_{BC}$ consists of Mean Squared Error (MSE) for the Cartesian ($\mathbf{xyz}$) and finger ($\mathbf{f}$) actions, and the unsigned relative rotation angle for the quaternion orientation ($\mathbf{q}$) action:
\vspace*{-2mm}
\begin{multline*}
L_{BC}(a,\hat{a}) = MSE(a_{xyz/f},\hat{a}_{xyz/f}) + 2\arccos({\hat{a}_{q}}^Ta_{q}).
\end{multline*}

$L_{BC}$ is calculated for both actions produced from privileged and onboard information encoder latents, against the demonstrated action $\hat{a}$.

\vspace{1mm}
\subsubsubsection{Latent Alignment loss.} $L_{align}$ is used to align the unit vector outputs of the privileged and onboard information encoders. We empirically find that Huber loss~\cite{huber1992robust} works better than MSE loss for the alignment on our tasks.

\vspace{1mm}
\subsubsubsection{Total loss.}
Overall, single-stage privileged training loss looks as follows:
\vspace*{-2mm}
\begin{multline*}
L_{total} = L_{BC}(\coloractR,\mathbf{\hat{a}}) + L_{align}(\colorzR,\colorzP) + L_{BC}(\coloractP,\mathbf{\hat{a}}).
\end{multline*}

\subsubsubsection{Training regimes.}
The combination of all loss terms simulates a single-stage privileged co-training regime, where:
\vspace{1mm}
\begin{enumerate}
    \setlength\itemsep{5pt}
	\item $L_{BC}(\coloractP,\mathbf{\hat{a}})$ trains the privileged \policyP~(a.k.a. \textit{Stage I})
	\item $L_{align}(\colorzR,\colorzP)$ emulates the transfer (a.k.a. \textit{Stage II})
	\item $L_{BC}(\coloractR,\mathbf{\hat{a}})$ adapts the behavior of  \policyR~for $\colorobs$.
\end{enumerate}

\vspace{1mm}
Through hyper-parameter search, we find that weighting all of the loss terms equally leads to the most efficient training regime of the onboard policy \policyR.

We also explore the variations of $L_{align}(\mathbf{sg(\colorzR}),\colorzP)$ and $L_{align}(\colorzR,\mathbf{sg(\colorzP}))$, to only allow for alignment in one direction by stopping the gradients ($\mathbf{sg()}$), however, we find no significant difference in the performance of \policyR.

\subsubsectioncol{Regularization} To regularize the network and make it robust to occlusions from different robot morphologies, we use image augmentation during the training process. We use the \textit{imgaug} library~\cite{imgaug} to augment camera images during training. For each image, we apply either \textit{Cutout} (random patch dropout) or \textit{CoarseDropout} (smaller but more frequent patch dropout) augmentation with a $50$\% chance.
We find these augmentations to be of high importance for training all our policies.

\subsubsectioncol{Training Parameters}
For all experiments in this paper we use the Adam optimizer with a learning rate of $3e-4$.
The total training time of the \shortname~policy \policy~takes five~days on a single V100 GPU.
\section{HARDWARE OPTIMIZATION}\label{sec:hw_optim}

\begin{figure}
  \centering
  \includegraphics[width=0.48\textwidth]{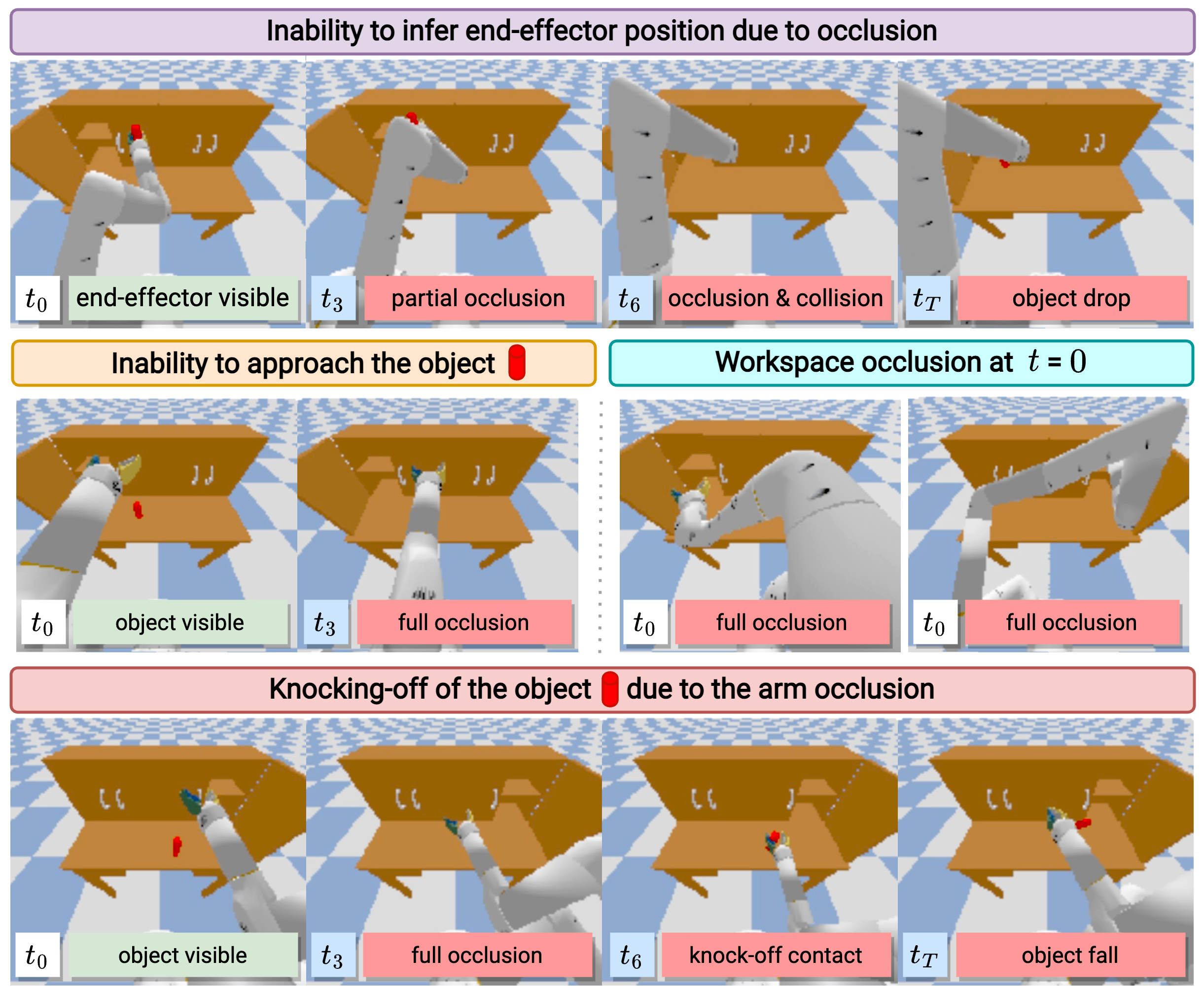}
  \caption{\textbf{Visual Occlusion Examples} - a selected set of naturally occurring object and workspace occlusions due to the pose and configuration of the robot.}
  \label{fig:occlusions}
\end{figure}

The goal of the hardware optimization is to find a robot morphology that is best fitted for the tasks and workspaces considered.
Unlike typical two-loop morphology optimization frameworks, our key idea is to leverage the morphology-agnostic policy \policyR\ trained only once.
Note that, policy \policyR\ through learning is \textit{primed} to disregard occlusions from onboard observation to generate suitable actions.
Hence, if the \policyR~is successful, then the information is sufficient, meaning that the morphology does not cause any major occlusions that prevent the task success (Fig. \ref{fig:occlusions} shows common types of occlusions).
Such an approach provides a more accurate evaluation of observability and reachability in terms of ``learnability'', and it is substantially more feasible than per-design training.
\newpage
\subsection{Objective Function}
\looseness=-1
The objective function for the optimization is the success rate of the morphology-agnostic policy \policyR~on the task of reaching and/or manipulation.
The success rate of \policyR~serves as a \textit{surrogate} measure of the robot morphology quality.
Each morphology is evaluated $500$ times on a seeded set of reaching tasks, and $200$ times on each of the $3$ manipulation tasks.

\subsection{Optimization Algorithm}
\looseness=-1
We use a black-box optimization algorithm, Vizier~\cite{google_vizier}, to optimize robot design parameters.
Vizier is a sampling-based optimization algorithm that allows us to efficiently search for the optimal solution among a large number of possible morphologies.
We evaluate $50$ morphologies in parallel capping the total number of evaluations at $1000$. 

\subsection{Optimization Space}
\looseness=-1
As an initial design configuration, we use Everyday Robots' manipulator robot~\cite{jang2022bc, saycan2022arxiv, rt12022arxiv}, which features a seven degree-of-freedom arm and a two-finger gripper.
We use this design as a baseline design and refer to it as a \textit{human-expert} design in the upcoming sections.

\looseness=-1
The optimizer is configured to propose link length deltas to the initial robot configuration. We allow for modification of the following robot links for respective ranges (in meters):
\textit{torso}~($-0.3,0.3$),
\textit{shoulder}~($0.0,0.3$),
\textit{bicep}~($-0.05,0.4$),\linebreak
\textit{elbow}~($-0.05,0.3$),
\textit{forearm}~($-0.2,0.3$),
\textit{wrist}~($0.0,0.2$),\linebreak
\textit{gripper}~($-0.05,0.2$).
We show a few random robot morphology configurations on the left of Fig. \ref{fig:architecture}.

\section{EXPERIMENTAL SETUP}\label{sec:exp_setup}

\looseness=-1
In this section, we describe the tasks, environment, and data collection process used for training the policy~\policy\ and optimization of the robot morphology. We also describe the training process of the targeted policy, which serves as a true performance evaluator for a given robot morphology.

\begin{figure}
  \centering
  \includegraphics[width=0.49\textwidth]{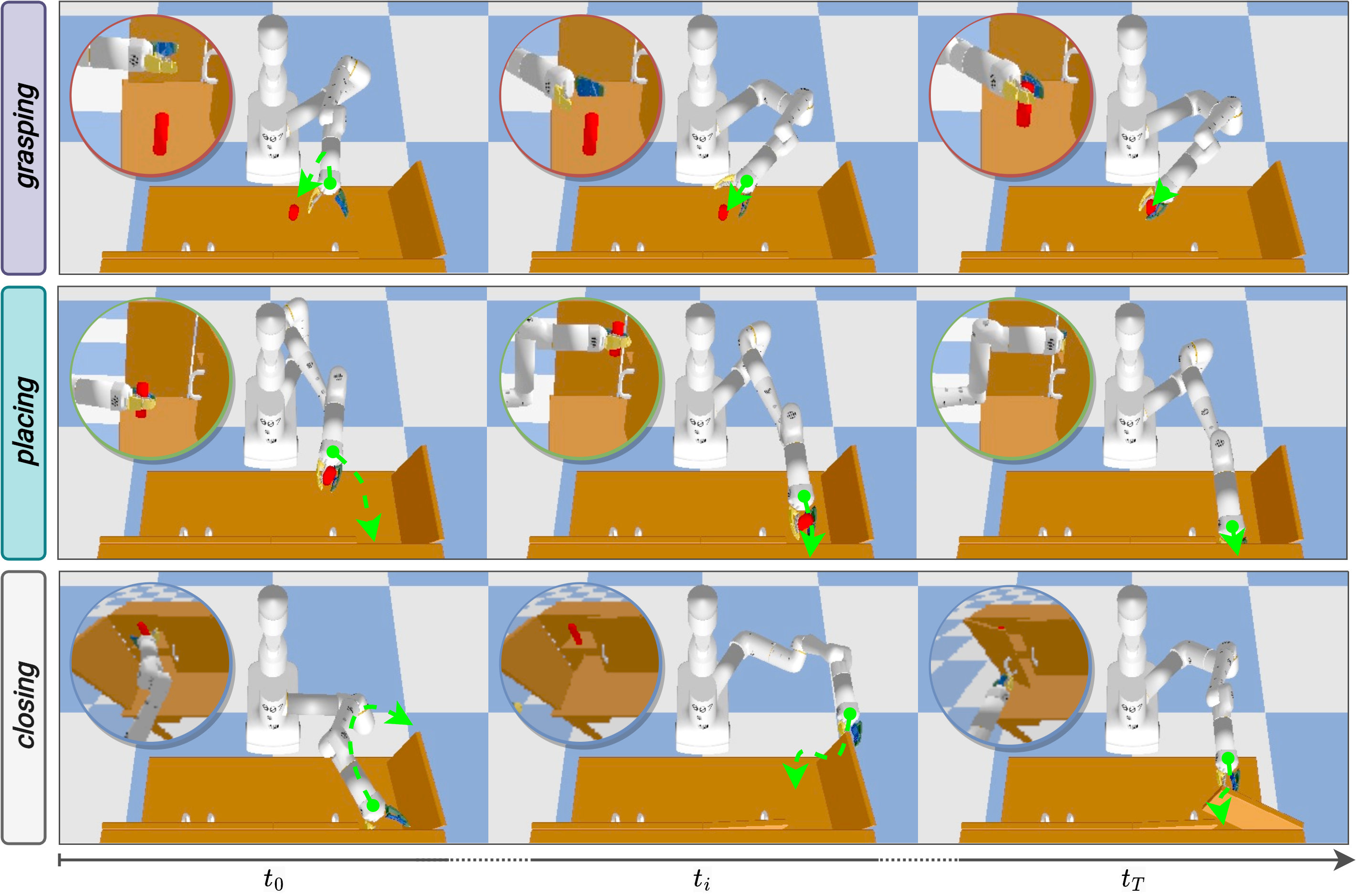}
  \vspace{-8mm}
  \caption{\textbf{Manipulation Task} - Robot motion key-frame visualization for object grasping, object placing, and cabinet closing sub-tasks.}
  \label{fig:task_motion_strip}
  \vspace{-5mm}
\end{figure}

\subsection{Tasks}
\looseness=-1
We use two tasks for training the policy: reaching and manipulation. The reaching task involves moving the end-effector to a specific goal position, while the manipulation task involves picking an object, placing it in a cabinet, and closing the cabinet door (Fig. \ref{fig:task_motion_strip}).

\subsubsectioncol{Reaching task}
\looseness=-1
The reaching task involves moving the robot's end-effector to a specific goal position within the workspace.
The goal position is randomly sampled at the beginning of each trial, and the robot's joint positions are also randomly placed in a different initial configuration.
The robot's end-effector is required to finish in a fixed orientation used across all trials.
The task is considered successful if the end-effector is within 3cm and 10 degrees from the goal pose for more than 20 time steps.
This task is used to evaluate the robot's ability to reach specific points within the workspace and helps test the onboard policies robot's arm/end-effector pose inference capabilities.

\subsubsectioncol{Manipulation task}
\looseness=-1
The manipulation task is a harder task, which is broken down into three sub-tasks: picking, placing, and closing.
The goal of this task is to evaluate the robot's ability to perform more complex manipulation actions, such as picking, manipulating objects, and interacting with the environment.
Unlike reaching, this task additionally requires that the network infers the object position and the cabinet states.

\subsubsubsection{Picking.}
\looseness=-1
For this task, the robot is required to pick up an object off the table surface.
The position of the object is randomized at the beginning of each trial, and the robot arm is reset into a randomized location in near proximity to the object.
The task is considered successful if the object is raised 20 cm above the table for more than 20 time steps.

\subsubsubsection{Placing.}
\looseness=-1
The robot must place the object in one of two cabinets. One of the cabinets is randomly selected as the goal at the beginning of each trial with its door being set as open.
The robot arm is initialized in the object-holding pose, with the goal to bring and release that object inside the target cabinet.

\begin{figure*}[!ht]
  \centering
  \begin{minipage}[b]{0.32\linewidth}
    \centering
    \includegraphics[height=6.5cm]{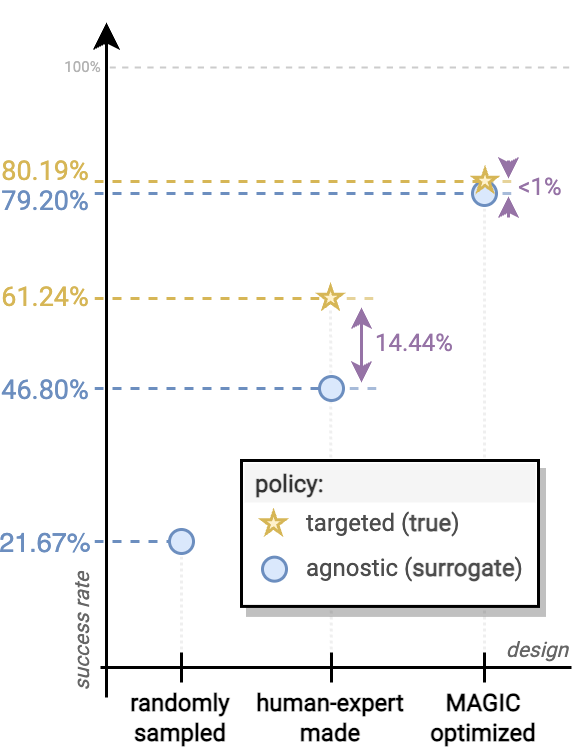}
    \caption{\textbf{Targeted vs. Morphology-Agnostic Policies.}
    Policies \policyR\ and \targetedpolicy\ have a $14.44\%$ performance gap on the \textit{human-expert} design, which shrinks to $1\%$ on the \ourmorph\ design.
    We omit the targeted performance evaluation on random morphologies due to computational cost.
    }
    \label{fig:optimized_vs_human}
  \end{minipage}
  \hfill
  \begin{minipage}[b]{0.35\linewidth}
    \centering
    \includegraphics[height=6.5cm]{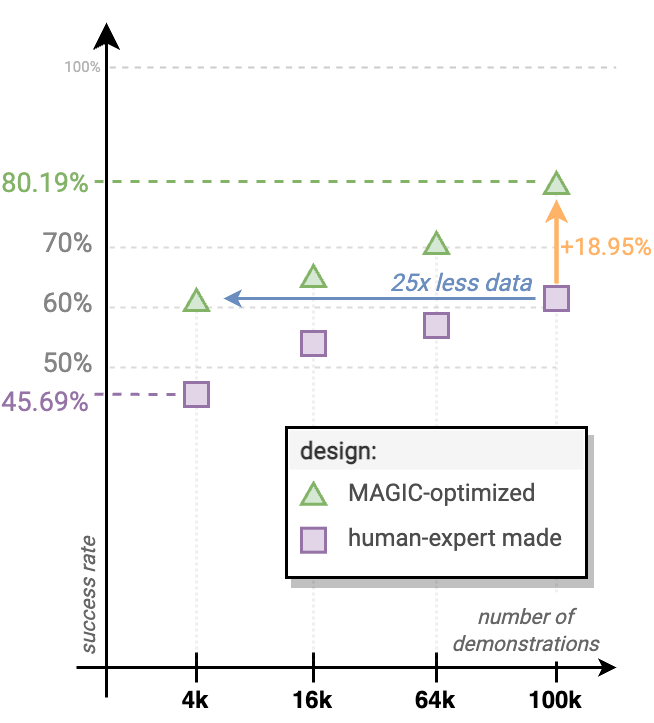}
    \caption{\textbf{Robot Performance \& Learning Efficiency.}
    \textbf{(a)} \textit{Learning oriented} morphology design outperforms the \textit{human-expert} design by $18.95\%$ when trained with the same amount of data.
    \textbf{(b)} \textit{Learning oriented} design can achieve the same performance as \textit{human-expert} design using $25$x less data.
    }
    \label{fig:better_learner}
  \end{minipage}
  \hfill
  \begin{minipage}[b]{0.29\linewidth}
    \centering
    \includegraphics[height=6.5cm]{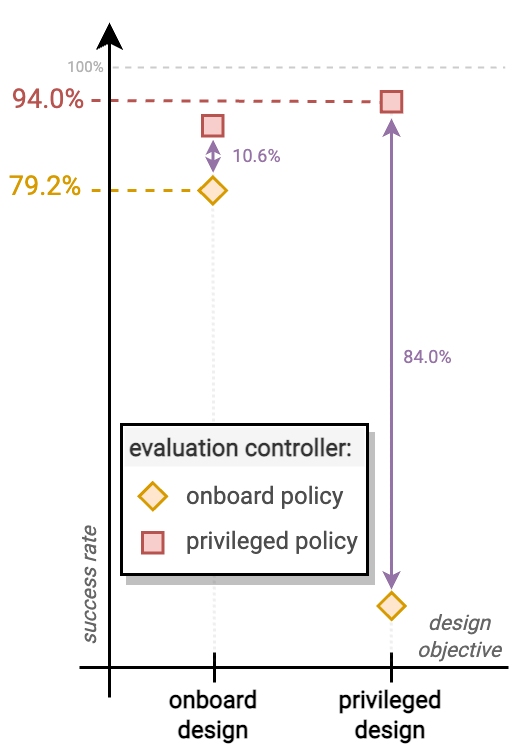}
    \caption{\textbf{Onboard and privileged design objective comparison.}  
    Morphology optimized using privileged objective has a major performance drop when evaluated using onboard information, emphasizing the importance of evaluation with onboard sensing.
    }
    \label{fig:best_priv_vs_best_real}
  \end{minipage}
  \label{fig:figures}
\end{figure*}

\subsubsubsection{Closing.}
\looseness=-1
The robot must close the door of the cabinet that the object was placed in.
The goal is to retract the arm from the post object placing pose and fully close the door of the cabinet.
The task is considered successful if the door is closed with a threshold of five degrees.

\subsection{Robot Data Collection}\label{subsec:data_collection}
\looseness=-1
Next, we describe the robot control and data collection procedures used for training the policy \policy\ and optimizing the robot morphology.
We collect trajectories using motion planning, which uses ground-truth simulation information, for various robot designs.
During training (Section \ref{sec:universal_pi}), \policyR\ is trained to produce these motions using only onboard observations.

\subsubsectioncol{Robot Control}
\looseness=-1
For our tasks, we use a robot manipulator with a seven degrees of freedom arm and a fixed base.
The robot arm is actuated through end-effector displacement control and Inverse Kinematics (IK) with collision avoidance.
During data collection, we use a set of predefined end-effector poses, which, IK guides the robot arm through avoiding any collisions.

\subsubsectioncol{Data Collection}
\looseness=-1
The data collection process involves initializing a morphology with uniformly sampled link lengths and collecting synthetic trajectories using the steps outlined in Algorithm \ref{alg:collection}.

\begin{algorithm}[H]
\caption{Cross-morphology data collection}\label{alg:collection}
\begin{algorithmic}[1]
\State Initialize buffer $\mathcal{B} \gets \{\}$

\While{$\mathcal{B}.size() < N$}
    \State Sample $m_i \sim \mathcal{M}$ \Comment{Uniformly Sample Morphology}
    \State workspace.reset() \Comment{Randomize Task/Workspace}
    \State plan $\gets$ planner(workspace, $m_i$) \Comment{Plan the Motion}
    \State $s_{0...T}$ $\gets$ execute(plan, workspace, $m_i$) \Comment{Execute}
    \If{$s_{T}.is\_success$} \Comment{Filter out failures}
        \State Store $s_{0...T}$ in $\mathcal{B}$
    \EndIf
\EndWhile
\State Return $\mathcal{B}$
\end{algorithmic}
\end{algorithm}
\vspace{-3mm}
All of the sub-tasks are timed, and the trajectory lengths are capped at 200 time steps ($\sim$10 seconds).
The data is collected using a planner which follows through a designated set of end-effector poses.
We use PyBullet~\cite{coumans2021} simulator to collect the demonstration data.
The data collection process is conducted in parallel by hundreds of workers to efficiently collect a large amount of data.
In total, we collect $N=500$k successful trajectories for training the policy \policy.

\subsection{Targeted Policy}\label{subsec:targeted}
A targeted policy \targetedpolicy\ is a controller designed to control a specific robot, unlike a \shortname~policy \policy, which is designed to perform well across multiple robots.
We use targeted policy as an ablation to evaluate how much the performance of our proposed \shortname~policy \policy is compromised to accommodate for multi-morphology capability.

To evaluate the true performance without multi-morphology compromise, we train the \textit{targeted} policy~\targetedpolicy\ using the procedure described in Section \ref{sec:universal_pi} but with a fixed robot design.
For a particular robot, we collect $100$k successful targeted demonstrations while keeping morphology~$m_i$ intact (skip Algorithm \ref{alg:collection}:\textit{line} 3) during the data collection process.
We use targeted demonstration to train the targeted policy~$\pi^T$ from scratch until convergence.
\section{EXPERIMENTS}
The results of our experimentation are presented in this section and structured to answer the following questions:

\begin{enumerate}
    \item[A.] Does a morphology-agnostic policy provide a good surrogate of a true performance?
    \item[B.] How does the \ourmorph~ design compare to a human-expert design?
    \item[C.] How does incorporating onboard sensing affects the morphology optimization process?
    \item[D.] What effects do additional robot tasks have on the optimized robot morphology?
\end{enumerate}

\subsection{Agnostic vs Targeted Policies}
First, we show how a surrogate policy performance compares to a true performance on a set of selected robot designs (see Fig.~ \ref{fig:optimized_vs_human}).
We report the performance of the morphology-agnostic controller \policyR\ (Section. \ref{sec:universal_pi}) and targeted controller~\targetedpolicy~(Section.~\ref{subsec:targeted}) on a \textit{human-expert} robot design~\mhuman\ and the best \ourmorph~solution candidate~\mours\ (described in Section~\ref{sec:hw_optim}).

We find that the reaching task performance of the \policyR\ and \targetedpolicy\ is nearly identical for \mours~(difference: $<1\%$).
On the other hand, the performance on \mhuman~design has a gap of $14.44\%$.
We hypothesize that \mours\ causes less occlusion and hence is easier to control using \policyR\ compared to \mhuman.
However, policy \targetedpolicy\; can learn to exploit the structure of occluding arm to infer the missing information, such as a rough pose of an end-effector, through targeted re-training.

\subsection{Comparison to a human-expert design}\label{sec:our_vs_baseline}
\looseness=-1
Next, we analyze the difference between \ourmorph~\mours\ and \textit{human-expert} \mhuman\ designs in terms of task performance and learning efficiency (Fig. \ref{fig:better_learner}).
We highlight the absolute performance improvement of \textit{targeted} policies when evaluated on \mours\ compared to \mhuman\ on the task of reaching.
Specifically, \mours\ achieves success rate of~$80.19$\% compared to~$61.24$\% with \mhuman.

\looseness=-1
The performance gain on \mours\ could be attributed to a reduced frequency of sensor occlusions.
We hypothesize that the reduced distortion of onboard information can also result in a higher quality of data during collection, which naturally leads to improved robot learning capabilities.
To measure the learning efficiency, we compare the success rates of the policies trained with a varying number of demonstrations.
Fig. \ref{fig:better_learner} shows that \mours\ is a better-suited robot for learning compared to~\mhuman, as it requires $25$x less data than \mhuman\ to reach the same performance when training the controller from scratch.

\subsection{Significance of onboard sensing}
\looseness=-1
Next, we seek to investigate the significance of onboard sensing during design optimization.
If the robot policy is given access to all of the information present in the environment, a robot may be able to reach its upper-bound performance, which is mostly constrained by kinematic and dynamic capabilities.
However, it might perform suboptimally when tested with onboard sensing because the morphology can limit its sensing capabilities via visual occlusions.

\looseness=-1
We use a privileged controller \policyP\ during the morphology optimization phase, which acts as an optimal motion planner with the ground-truth robot and task information.
Once the morphology is found, we evaluate its performance with the policy which relies on onboard information \policyR.
In Fig.~\ref{fig:best_priv_vs_best_real}, we compare the performance of privileged information morphology and onboard sensing morphology \mours.
We observe that if we only use the privileged policy during the morphology design, there is a significant $84\%$ drop in performance when the robot is evaluated with onboard sensing. In contrast, \mours\ performs well in both onboard and privileged settings.
This result demonstrates the significance of using onboard sensing during the robot design process.

\subsection{Multitask-based regularization}

\begin{figure}
  \centering
  \vspace{3mm}
  \includegraphics[width=0.49\textwidth]{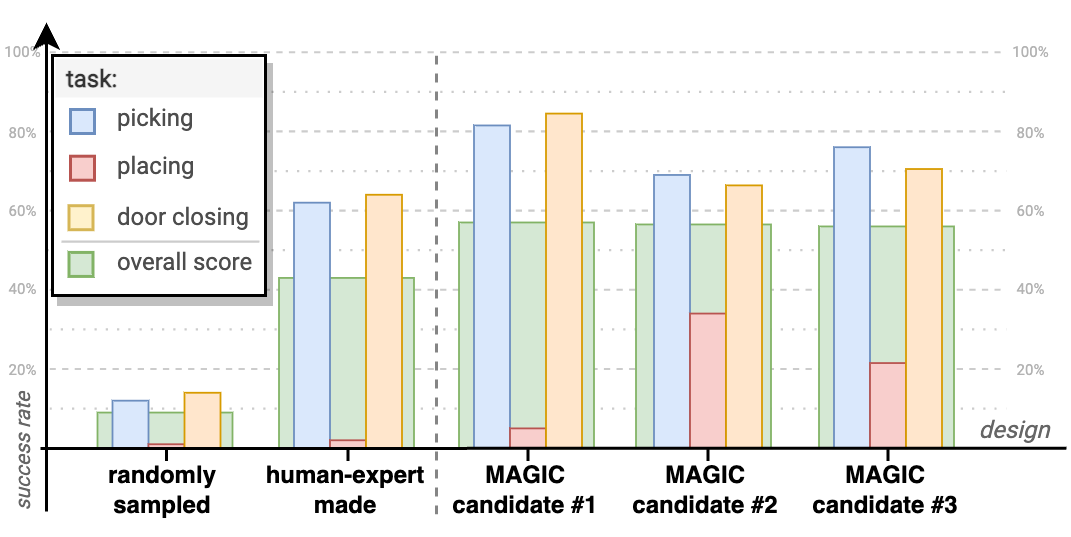}
  \caption{\textbf{Various design performance comparison on the manipulation task}.
  Overall \ourmorph\ top solution candidates perform $\sim15\%$ better than \textit{human-expert} design.
  Different optimization candidates show varying levels of performance at different sub-tasks, allowing a human to make a final decision about the performance trade-offs.
  }
  \label{fig:manipulation_stats}
  \vspace{-5mm}
\end{figure}

\begin{figure}
  \centering
  \vspace{10mm}
  \includegraphics[width=0.48\textwidth]{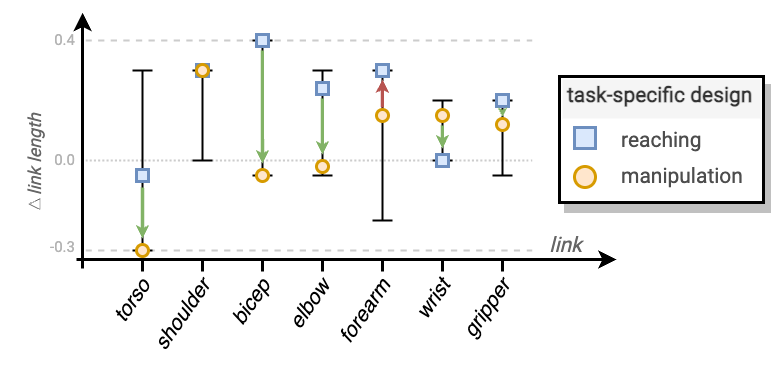}
  \vspace{-8.5mm}
  \caption{\textbf{Task Complexity Effects on Robot Design} - 
  Optimized link lengths of the robot morphologies tend to converge to more plausible solutions when optimized on more complex tasks.
  }
  \label{fig:morph_regularization}
  \vspace{1mm}
\end{figure}

\looseness=-1
Finally, we investigate the direction of task-based regularization, through the exposure of the robot to a wider set of manipulation tasks.
There exists a large number of regularization approaches applied during the optimization to improve the practicality of the design such as energy, or material penalties~\cite{coello1998using, kouritem2022multi}.
However, we intend to avoid an explicit regularization that can directly impact the final morphology. Instead, we seek implicit regularization via learning on multiple tasks.

We repeat data collection, training, and optimization procedures with manipulation tasks and report the performance of multiple solution candidates in Fig. \ref{fig:manipulation_stats}.
In addition to seeing similar trends in performance improvements of \ourmorph~design \mours\ in manipulation similar to reaching task, we also observe overall improvements in the robot design solutions.
Fig.~\ref{fig:morph_regularization} compares two robot morphologies: one optimized for the reaching task, and one that is optimized for the manipulation task.
Not surprisingly, the type and complexity of the tasks being considered could significantly impact the optimal morphology design.
Hence, including multiple tasks during the optimization process can lead to a more regularized morphology, with shorter and more manageable links that are likely easier to manufacture.
\section{CONCLUSION}
\looseness=-1
In this work, we explore the potential of optimizing robot morphology for improved learning and demonstrate that it can be achieved through a holistic task-oriented optimization process.
We propose a cost-effective method to improve the robot hardware by training a morphology-agnostic surrogate controller and demonstrate that a robot designed with learning considerations can excel at learning compared to a human-expert design.
We introduce a single-stage privileged learning framework that enables rapid acquisition of an onboard policy without artifacts on two-stage privileged learning transfer.
Through our experiments, we show that robot designs with enhanced learning capabilities can improve performance by $15\%$ on manipulation and by $20\%$ on complex manipulation tasks compared to a human-expert robot design.
Moreover, an optimized robot reaches a human-expert design performance level with $25$x less demonstration data. 
We hope this work contributes to the growing trend of learning-based robots and sheds light on opportunities in hardware designs that facilitate better learning.
\newpage
\section{Future Work}

While using simulation was necessary for making it tractable to evaluate hundreds of different morphologies, a natural next step is to build the final optimized robot design in the real world and test its performance against the simulation results.
Another extension of our work is to include sensor placement during the optimization as well as evaluating on more complex task distributions.
Investigations in those directions could help us discover some unconventional designs that are overlooked by human engineers.

\bibliographystyle{IEEEtran}
\bibliography{refs}
\end{document}